# A Literature Survey of various Fingerprint De-noising Techniques to justify the need of a new De-noising model based upon Pixel Component Analysis

Dr Siddharth Choubey
Shri Shankaracharya Group of Institutions, Bhilai
Email id-sidd25876@gmail.com

Deepika Banchhor
Shri Shankaracharya Group of Institutions, Bhilai
Email id-deepika.banchhor@gmx.us

**ABSTRACT:**

Image Preprocessing is a vital step in the field of image processing for biometric pattern recognition. This paper studies and reviews various classical and modern fingerprint image de-noising models. The various model used for de-noising ranges widely from transform matrix using frequency, histogram model de-noising, de-noising by introducing Gabor filter and its types to enhance fingerprint images.

The output efficiency of various de-noising model proposed earlier is calculated on the basis of SNR (signal to noise ratio) and MSE (mean square error rate). Our simulated experimental results indicates that incorporating the de-noising model based on Gabor filter inside domain of wavelet ranges with composite method only betters MSE (Mean Square Error). Improved MSE without significant improvement in SNR improves the fingerprint images only by a little margin which is non-optimal in nature. Thus the objective of this research paper is to build an optimal de-noising model for fingerprint images so that its usage in biometric authentication can be more robust in nature.

**INTRODUCTION:**

Fingerprints are used extensively for biometric applications due of their exclusive characteristics of uniqueness and similarity over time. The ability to record unique characteristics in different portions of the finger lines is the main attribute of fingerprint, which can be useful in a number of applications. However, thermal effects, sensor saturation, quantization errors, and dirt generate a noise that deteriorates the quality and are also affected by multiplicative noise in addition to additive noise; thereby creates a bad effect on image analysis. Thus original fingerprint imagery is prone to different types of noises. Some of these noises incorporate useless characteristics, which are faulty considerations for the calculations of quintessential minutia of the fingerprint. The existing technologies of fingerprint de-noising have yet not achieved the optimal level of de-noising. To tend to these issues, a new de-noising technique is required for robust and efficient rendering of fingerprint images. Thus in this research paper through survey we are exploring a new technique of fingerprint image de-noising based on pixel component analysis (PCA). The proposed de-noising model gives improved results when compared to the existing models. Initial results indicate that the proposed de-noising algorithm gives optimal de-noised fingerprint image for practical applications.

Fingerprint as means of biometric verification is steadily growing in popularity as a digital means for authentication in both commercial and personal spheres. It is frequently used as a substitute when thumb impression, signature and other classical measures are outdated or inadequate. The ability to record unique characteristics in different portions of the finger lines is the main attribute of fingerprint, which can be useful in a number of applications.

Fingerprints are widely used in a variety of security applications .Fingerprints are unique to each person and do not change overtime. Hence it is considered to be a more reliable approach in solving personal identification problems.

Fingerprints are similar to graphical patterns present on human fingers as lines which may be termed as ridges. A ridge is simply a single line curve and a valley is the space amongst two nearest ridges. A fingerprint image biometric verification system consists of four operations namely - ridge optimization for authentication, ridge characteristic evaluation, segmentation and matching with the pre-existing standard authorized fingerprint local database. The acquisition of a fingerprint image can be done either by scanning an inked expression or through live electronic biometric scanner. The fingerprint image must have optimum quality for biometric authentication and recognition system. The common problems linked with fingerprint images are discontinuous ridges, i.e., gaps or island formation, improper separation of ridges, cuts and bruises, dirt, etc.

**CURRENT FINGERPRINT DE-NOISING MODELS**

There are several de-noising models for fingerprint images have been proposed till today. Some of the important fingerprint image de-noising models are Gabor Filters with FFT (Fast Fourier Transform) based FPI de-noising model, Fusion and Context Switching frameworks fingerprint matching mechanism, segmentation algorithm based FPI de-noising, Gaussian de-noising, histogram based processing and cancelling of FPI noise, etc.

Gabor filters are tuned to the local ridge curve angles and frequency. This algorithm includes preprocessing normalization, calculation of ridge orientation and frequency and then filtering. But the curve angles are often erroneously calculated due to data leakage while applying Gabor filters. Hence there is a significant short coming in this method of FPI de-noising.

Fourier filtering uses the special co-ordinates of a fingerprint image to de-noise. But this process involves more time taxing operations such as the pre-filtering, estimation of 16 directions based orientation, etc. Also the de-noised image is not optimal.

Visu shrink is another method of fingerprint de-noising. It is a hard threshold method where the threshold value is kept in ratio with the noise's standard deviation. It also fails to lessen mean square error (MSE). Hence it is not optimal in nature.

Overall, it is observed that most models for fingerprint image de-noising are based upon filters that are varying according to the global and local characteristics of FPI (fingerprint images). Ridge curve angles and valleys seem to be the most prominent characteristics of the fingerprint image which are analyzed and enhanced for de-noising purposes. But due to several limitation these methods of fingerprint de-noising are non optimal in nature and requires further improvement.

**PIXEL COMPONENT ANALYSIS**

We are proposing an independent algorithm based on pixel component analysis. The pixel component analysis utilizes the pixel characteristics and each individual pixel analysis check for homogeneity thus replacing all the non-homogenous pixels with patches from neighboring pixels to maintain homogeneity and thus eliminating noise.

**PROPOSED STEPS**

1. A fingerprint image is taken. The acquisition can either be done via impression or direct scanning of the fingerprint.
2. The fingerprint image vitals such as MSE (Mean Square Error) and SNR (Signal Noise Ratio) is estimated.
3. The image is preprocessed to fit console window size and converted into grayscale format.
4. The FPI pixels rows and columns are calculated.
5. The FPI pixels are divided into pixel clusters of 3x3.
6. Each pixel cluster will be segregated from the original FPI.
7. We will perform component analysis on each pixel cluster to check homogeneity of the cluster. Here non-homogeneity indicates the noise.
8. If the pixel cluster is homogenous it is already optimal in nature. If it is non homogenous, then we will remove the non-homogenous pixel from that pixel cluster.
9. We will replace the non homogenous pixel with patches from neighboring pixels of that clusters.
10. Repeat step 7 to 9 for each pixel cluster.
11. We will place each de-noised pixel cluster in its relevant place in de-noised fingerprint image using normalization using threshold.
12. We will get the de-noised image.

**EXPECTED RESULT**

As each set of pixels are being de-noised in this process and the problem of Lossy compression and data leakage are non-existent, we believe that if performed practically, the proposed algorithm will definitely yield better result than the existing models of fingerprint image de-noising such as Gabor filters, Gaussian filter, Fourier transform, etc.

**CONCLUSION AND FUTURE SCOPE**

The proposed methodology seems to be better equipped for de-noising the fingerprint images than the classical methods. So we are going to adopt PCA (Pixel Component Analysis) method in our future research for fingerprint de-noising. The agglomeration of PCA (Pixel Component Analysis) with BRNN (Bi-Directional Recurrent Neural Network) is also likely to yield better results and can be worked in future. Hence the need of a new independent algorithm for fingerprint image de-noising is established.